\documentclass[10pt,conference,a4paper]{IEEEtran}

\usepackage{times}
\usepackage{epsfig}
\usepackage{graphicx}
\usepackage{amsmath}
\usepackage{amssymb}

\newcommand{\etal}{\textit{et al}.}

\ifCLASSINFOpdf
\else
\fi
\begin{document}
%
\title{GP-GAN: Gender Preserving GAN for Synthesizing\\ Faces from Landmarks}

\author{\parbox{16cm}{\centering
    {\large Xing Di, Vishwanath A. Sindagi, Vishal M. Patel }\\
    {\normalsize
    Rutgers, The State University of New Jersey \\
    94 Brett Road, Piscataway, NJ 08854}\\
    {\tt\small \{xing.di, vishwanath.sindagi, vishal.m.patel\}@rutgers.edu}}
}

%


\maketitle

\begin{abstract}
Facial landmarks constitute the most compressed representation of faces and are known to preserve information such as pose, gender and facial structure present in the faces. Several works exist that attempt to perform high-level face-related analysis tasks based on landmarks alone without the aid of face images.  In contrast, in this work,  an attempt is made to tackle the inverse problem of synthesizing faces from their respective landmarks. The primary aim of this work is to demonstrate that information preserved by landmarks (gender in particular) can be further accentuated by leveraging generative models to synthesize corresponding faces. Though the problem is particularly challenging due to its ill-posed nature,  we believe that successful synthesis will enable several applications such as boosting performance of high-level face related tasks using landmark points and performing dataset augmentation. To this end, a novel face-synthesis method known as Gender Preserving Generative Adversarial Network (GP-GAN) that is guided by adversarial loss, perceptual loss and a gender preserving loss is presented.  Further, we propose a novel generator sub-network UDeNet for GP-GAN that leverages advantages of U-Net and DenseNet architectures. Extensive experiments and comparison with recent methods are performed to verify the effectiveness of the proposed method. Our code is available at: https://github.com/DetionDX/GP-GAN-Gender-Preserving-GAN-for-Synthesizing-Faces-from-Landmarks 
\end{abstract}


%
\IEEEpeerreviewmaketitle

\section{INTRODUCTION}

Facial landmarks can be regarded as the most compressed representation of a face due to the fact that very few number of points are required to capture the landmark locations.  In spite of the incredibly low number of keypoints,  they are known to preserve important information about the face such as pose, gender \cite{cao2011can} and structure \cite{wu2012age,Qianru2018HeadInpainting,wei2018UniqueSmile}. Success of facial analysis tasks using just landmark keypoints is essential from the perspective of memory management and information privacy. Considering that size of landmarks is an order of magnitude smaller as compared to the image size, it will result in significant savings in terms of memory. Essentially, we are now able to store only landmark key points and throw away face image for a particular application. In addition, landmark information can be safely stored, transported, and distributed without potential violation of human privacy and confidentiality. Motivated by these reasons, it would be interesting to understand how landmarks can be exploited for performing high-level facial analysis tasks in the absence of corresponding face images. 

Several researchers have demonstrated that facial landmarks can be used in many face analysis tasks such as  face recognition \cite{JC_CNN,HyperFace,DeepFace_SPM2018}, facial attribute inference \cite{zhang2014facial}, age estimation \cite{wu2012age}, gender recognition \cite{cao2011can} and expression analysis \cite{taheri2011towards}. However, these methods operate on a small set of keypoints due to which their performance is severely limited. To overcome this problem, we propose a novel solution that involves synthesis of faces from landmark points using the recently popular generative models \cite{goodfelllow2014generative,zhu2017unpaired,berthelot2017began,zhao2016energy,zhang2017image,lidan2017highquality,zhang2018densely}. While, several methods \cite{zhang2014facial,valstar2010facial,kumar2016face,sun2013deep} have been proposed in the literature for landmark detection, the inverse problem of synthesizing faces from their corresponding landmarks is a largely unexplored problem.  We believe that using synthesized faces will result in better recognition performances as they leverage the capabilities of generative models to accentuate information present in landmarks.  Apart from their use in high-level facial analysis tasks, these generative methods can be used to create virtually unlimited stochastic samples by conditioning on both landmarks and a stochastic noise vector enabling us to augment existing datasets for large scale learning \cite{bousmalis2016unsupervised}. 

\begin{figure}[t!]
	\centering
	\includegraphics[width=0.8\linewidth]{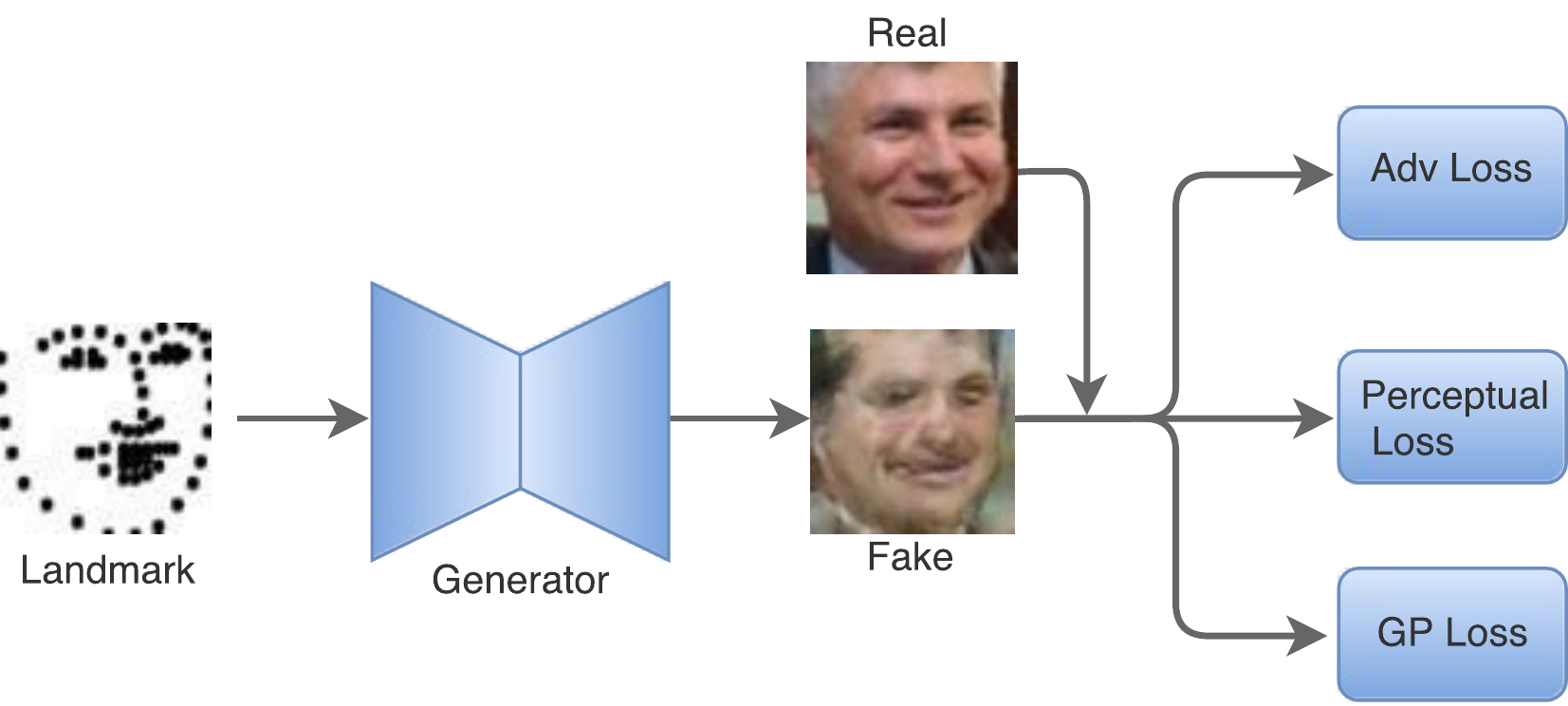}
	\vskip -8pt\caption{Overview of the proposed GP-GAN method for synthesizing faces from landmarks. In addition to adversarial loss function, the generator sub-network is guided by a perceptual loss and a gender preserving loss. }
	\label{fig:overview}
\end{figure}

In this work, generative models are exploited to synthesize faces from landmarks in an attempt to accentuate information (gender in particular) present in the landmarks. Cao \etal \cite{cao2011can} specifically address the question if facial metrology can be used to predict gender and they further go on to demonstrate that gender recognition using landmarks achieves reasonable performance. This is remarkable considering the fact that only 68 keypoints are used to predict gender of the face represented by these keypoints. However, generating faces from landmarks will enable us to achieve further improvement in  performance as this process will leverage generative models to learn the distribution of landmarks and their mappings to the respective faces. While recognition of other attributes like ethnicity, pose, identity, etc. can all be improved, in this work, we specifically focus on the gender attribute. To this end, we propose Gender Preserving Generative Adversarial Network (GP-GAN) to generate faces from their respective landmarks (as shown in Fig. \ref{fig:overview}). To further enhance the network's performance, it is guided by perceptual loss and a gender preserving loss in addition to adversarial loss. To summarize, following are the key contributions of this work:
\begin{itemize}
	\item To the best of our knowledge, this is the first attempt to generate faces from landmark keypoints while preserving gender information. 
	\item A GAN-based framework guided by perceptual and gender preserving loss is proposed. The generator is constructed using a novel combination of UNet \cite{ronneberger2015u} and DenseNet \cite{huang2017densely}, which we call it as UDeNet. 
	\item Detailed experiments are conducted to demonstrate the improvements in gender recognition obtained from synthesized images using the proposed method. 
\end{itemize}

\begin{figure*}[ht!]
	\centering
	\includegraphics[width=.5\linewidth]{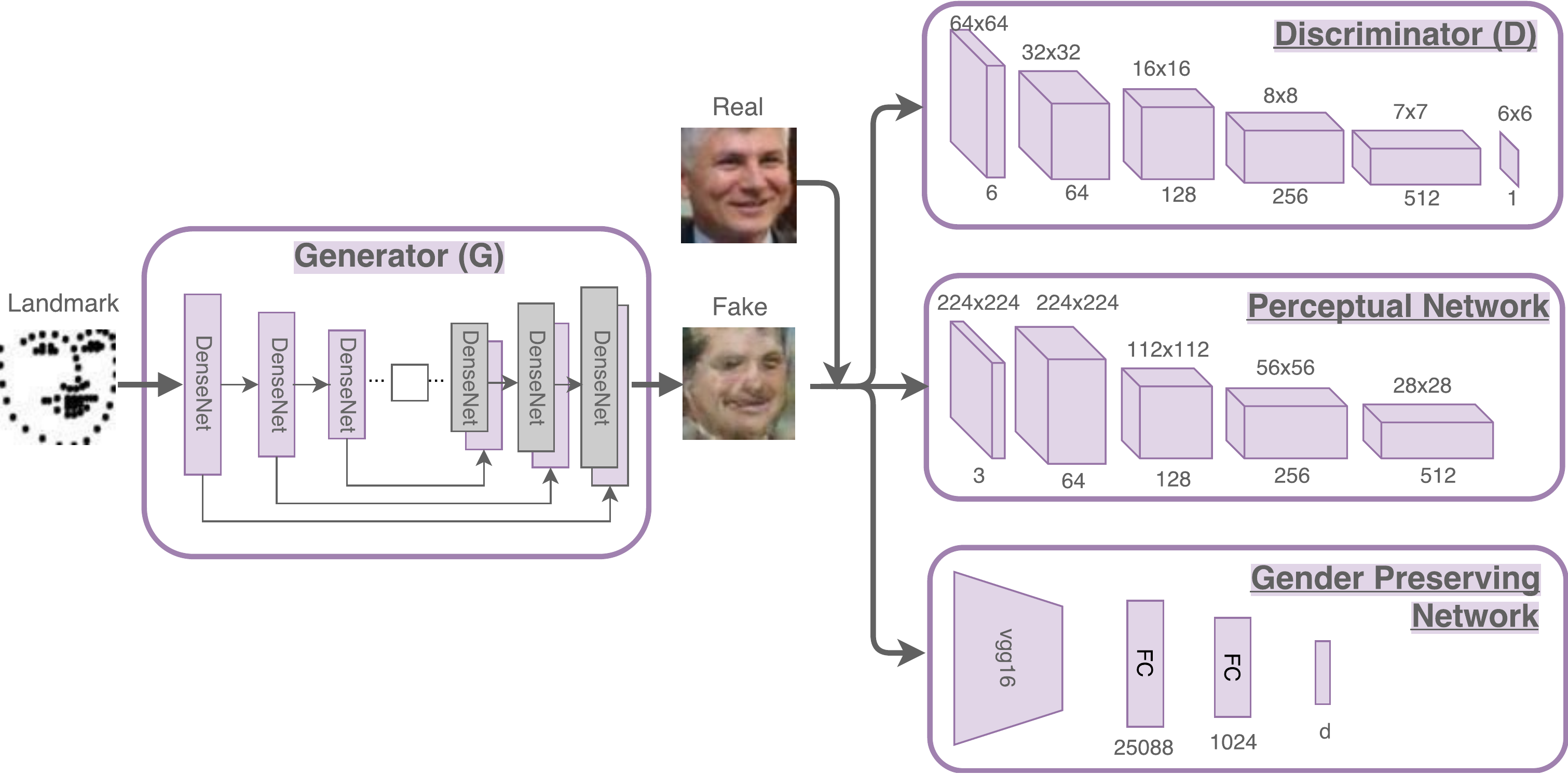}
	\hskip45pt
	\includegraphics[width=.15\linewidth]{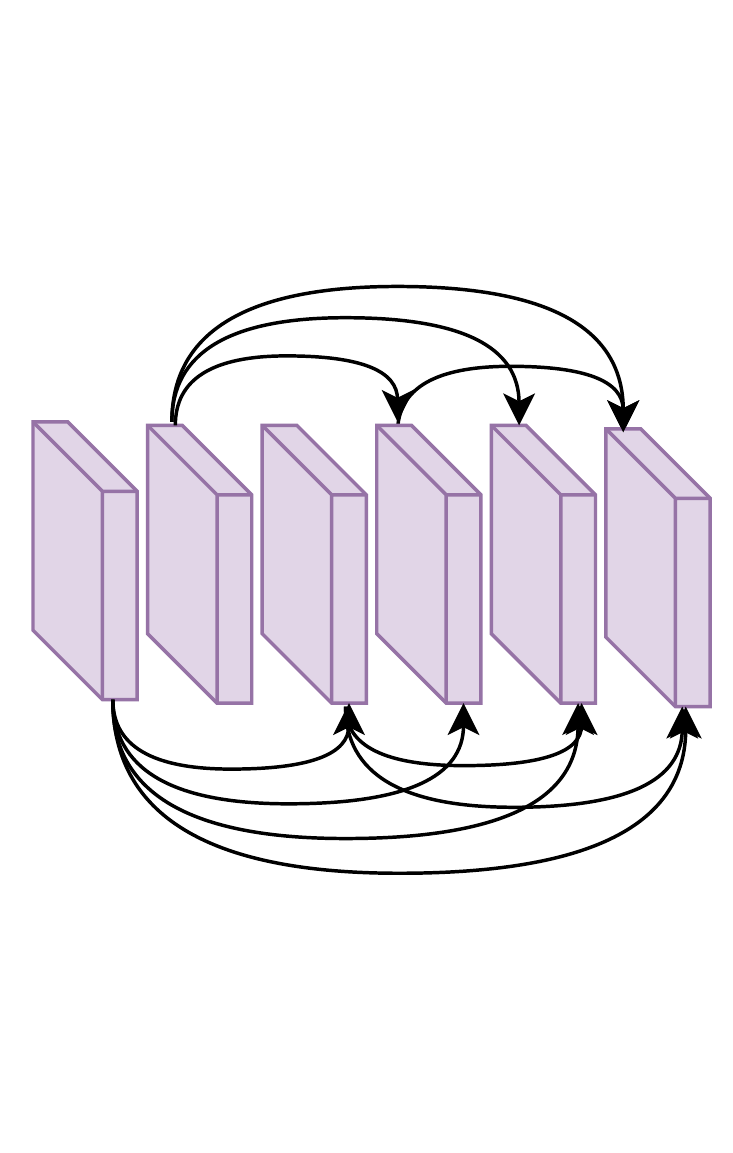}\\
	\vskip-8pt\caption{Architecture of the proposed GP-GAN framework. Left: Generator ($G$) synthesizes face image from landmarks and is based on UNet and DenseNet architecture. $D$ is a patch-based discriminator that is trained to distinguish between real/fake face images and it is responsible for providing adversarial feedback to $G$. $G$ is also guided by a perceptual loss (based on VGG-16 architecture) and a gender-preserving loss. Right: Dense-block used in generator $G$. }
	\label{fig:framework}
\end{figure*}

\section{RELATED WORK}


In contrast to landmark detection methods \cite{saragih2011deformable,sun2013deep,kumar2016face,asthana2013robust} , we focus on the inverse problem of synthesizing or generating faces from landmark keypoints which is a relatively unexplored problem. To this end, recently popular generative models are explored in this work. 
Among these methods, we specifically study Generative Adversarial Network (GAN)\cite{goodfelllow2014generative,zhu2017unpaired,berthelot2017began,zhao2016energy} and Variational Auto-encoder (VAE) \cite{rezende2014stochastic,kingma2013auto}.

VAEs are powerful generative models that use deep networks to describe distribution of observed and latent variables. A VAE consists of two networks, with one network encoding a data sample to a latent representation and the other network decoding  latent representation back to data space. VAE regularizes the encoder by imposing a prior over the latent distribution. Conditional VAE (CVAE) \cite{sohn2015learning} \cite{yan2016attribute2image} \cite{di2017face} is an extension  of VAE that models latent variables and data, both conditioned on side information such as a part or label of the image. GANs  \cite{goodfelllow2014generative} are another class of generative models that are used to synthesize realistic images by effectively  learning the distribution of training images. Recently, several variants based on this game theoretic approach have been proposed for image-to-image translation tasks. Isola \etal \cite{isola2016image} proposed Conditional GANs \cite{mirza2014conditional} for several tasks such as labels to street scenes, labels to facades, image colorization, etc. In an another variant, Zhu \etal \cite{zhu2017unpaired} proposed CycleGAN that learns image-to-image translation in an unsupervised fashion. Berthelot \etal \cite{berthelot2017began} proposed a new method for training auto-encoder based GANs that is relatively more stable. Their method is paired with a loss inspired by Wasserstein distance. Some of the other applications of GANs include image de-hazing \cite{zhang2017joint}, crowd counting \cite{sindagi2017generating}, and image de-raining \cite{zhang2017image}.

\section{Proposed Method}

Given an application where only facial landmarks are available, we explore how to leverage information preserved in these keypoints. To this end, we propose to model the joint distribution of facial landmarks and corresponding face images \footnote{Face images are available only during training} using generative modeling. Inspired by the success of GANs \cite{goodfelllow2014generative}, we explore adversarial networks in this work for synthesizing faces from landmark keypoints. GANs, motivated by game theory, consist of two competing networks: generator $G$ and discriminator $D$. The goal of GAN is to train $G$ to produce samples from training distribution such that the synthesized samples are indistinguishable from actual distribution by discriminator $D$. Conditional GAN is another variant where the generator is conditioned on additional variables such as discrete labels \cite{mirza2014conditional}, text \cite{reed2016generative} and images \cite{isola2016image}. 
The objective function of a conditional  GAN is defined as follows
\begin{equation}\label{eq:conditional GAN loss}
	\begin{split}
		L_{cGAN}(G,D) = E_{x,y \sim Pdata (x,y)}[\log D(x,y)]+ \\
		E_{x\sim Pdata(x),z\sim p_{z}(z)}[\log(1-D(x,G(x,z)))],
	\end{split}
\end{equation}
where $y$, the output image, and $x$, the observed image, are sampled from distribution $Pdata (x,y)$ and they are distinguished by the discriminator, $D$. While for the generated fake $G(x,z)$ sampled from distributions $x\sim Pdata(x),z\sim p_{z}(z)$ would like to fool $D$.

As shown in Fig. \ref{fig:framework}, the proposed network consists of a generator sub-network $G$ (based on U-net \cite{ronneberger2015u} and DenseNet \cite{huang2017densely} architecture) conditioned on a facial landmark image and a patch-based discriminator sub-network $D$. $G$ takes landmark as input and attempts to generate corresponding face image, while $D$ attempts to distinguish between real and synthesized images. The two sub-networks are trained iteratively. In addition to the adversarial loss, we propose to guide the generator using three other loss functions: perceptual loss based on VGG-16 architecture \cite{simonyan2014very}, gender preserving loss and $L_1$ reconstruction error.

\subsection{Generator}
Deeper networks are known to better capture high-level concepts, however, the vanishing gradient problem affects convergence rate as well as the quality of convergence. Several works have been developed to overcome this issue among which U-Net \cite{ronneberger2015u} and DenseNet \cite{huang2017densely} are of particular interest. While U-Net incorporates longer skip connections to preserve low-level features, DenseNet employs short range connections within micro-blocks resulting in maximum information flow between layers in addition to an efficient network. Motivated by these two methods, we propose UDeNet for the generator sub-network $G$ in which, the U-Net architecture is seamlessly integrated into the DenseNet network in order to leverage advantages of both the methods. This novel combination enables more efficient learning and improved convergence quality. 

A set of 3 dense-blocks (along with transition blocks) are stacked in the front, followed by a set of 5 dense-block layers (transition blocks). The initial set of dense-blocks are composed of 6 bottleneck layers. For efficient training and better convergence, symmetric skip connections are involved into the  generator sub-network, similar to \cite{mao2016image}. Details regarding the number of channels for each convolutional layer are as follows: \linebreak
 C(64)-M(64)-D(256)-T(128)-D(512)-T(256)-D(1024)-T(512)-D(1024)-DT(256)-D(512)-DT(128)-D(256)-DT(64)-D(64)-D(32)-D(32)-DT(16)-C(3), \linebreak 
 where C(K) is a set of $K$-channel convolutional layers followed by batch normalization and ReLU activation. M is max-pooling layer. D(K) is the dense-block layer with $K$-channel output, T(K) is transition layer with $K$-channel output for downsampling. DT(K) is similar to T(K) except for transposed convolutional layer instead of convolutional layer for upsampling. 

\subsection{Discriminator}

Motivated by \cite{isola2016image}, patch-based discriminator $D$ is used and it is trained iteratively along with $G$. The primary goal of $D$ is to learn to discriminate between real and synthesized samples. This information is backpropagated into $G$ so that it generates samples that are as realistic as possible. Additionally, patch-based discriminator ensures preserving of high-frequency details which are usually lost when only L\textsc{1} loss is used. All the convolutional layers in $D$ have a filter size of $4\times4$.  Details regarding the number of channels for each convolutional layer are specified in Fig. \ref{fig:framework}.

\subsection{Objective function}

The network parameters are learned by minimizing the following objective function:  
\begin{equation}\label{generator loss}
	L = L_{A} + \lambda_{P}L_{P} + \lambda_CL_C + \lambda_1L_1,
\end{equation}
where $L_{A}$ is the adversarial loss, $L_{P}$ is the perceptual loss, $L_{C}$ is the gender preserving loss and $L_{1}$ is the loss based on $L_1$-norm between the target and reconstructed image, $\lambda_{P}$, $\lambda_{C}$ and $\lambda_{1}$ are weights respectively for perceptual loss, gender preserving loss and $L_{1}$ loss. \\

\noindent\textbf{Adversarial loss:} Adversarial loss is based primarily on the discriminator sub-network $D$. Given a set of $N$ synthesized faces, $\{\hat{x}_i\}_{i=1}^{N}$, the entropy loss from $D$ that is used to learn the parameters of $G$ is defined as: 
\begin{equation} \label{eq:advloss}
	L_{A} = -\frac{1}{N}\sum_{i=1}^{N}log(D(\hat{x}_i)),
\end{equation} \\

\noindent\textbf{Perceptual loss:} Johnson \etal \cite{johnson2016perceptual} introduced  the perceptual loss function for style transfer and super-resolution. Instead of relying only on $L_1$ or $L_2$ reconstruction error, they learn the network  parameters using errors between high-level image feature representations extracted from a pre-trained convolutional neural network. Similar to their work, pre-trained VGG-16 \cite{simonyan2014very} network is used to extract high-level features (conv4\_3 layers) and the $L_1$ distance between these features of real and fake images is used to guide the generator $G$. The perceptual loss function is defined as: 

\begin{equation} \label{eq:perceptual loss}
	L_{P} = ||V(\hat{x})-V(x)||_{1},
\end{equation}
where,  $x$ and $\hat{x}$ indicate real and fake images, respectively and $V$ is a particular layer of the VGG-16 network. \\

\begin{figure}[ht!]
	\centering
	\includegraphics[width=1\linewidth]{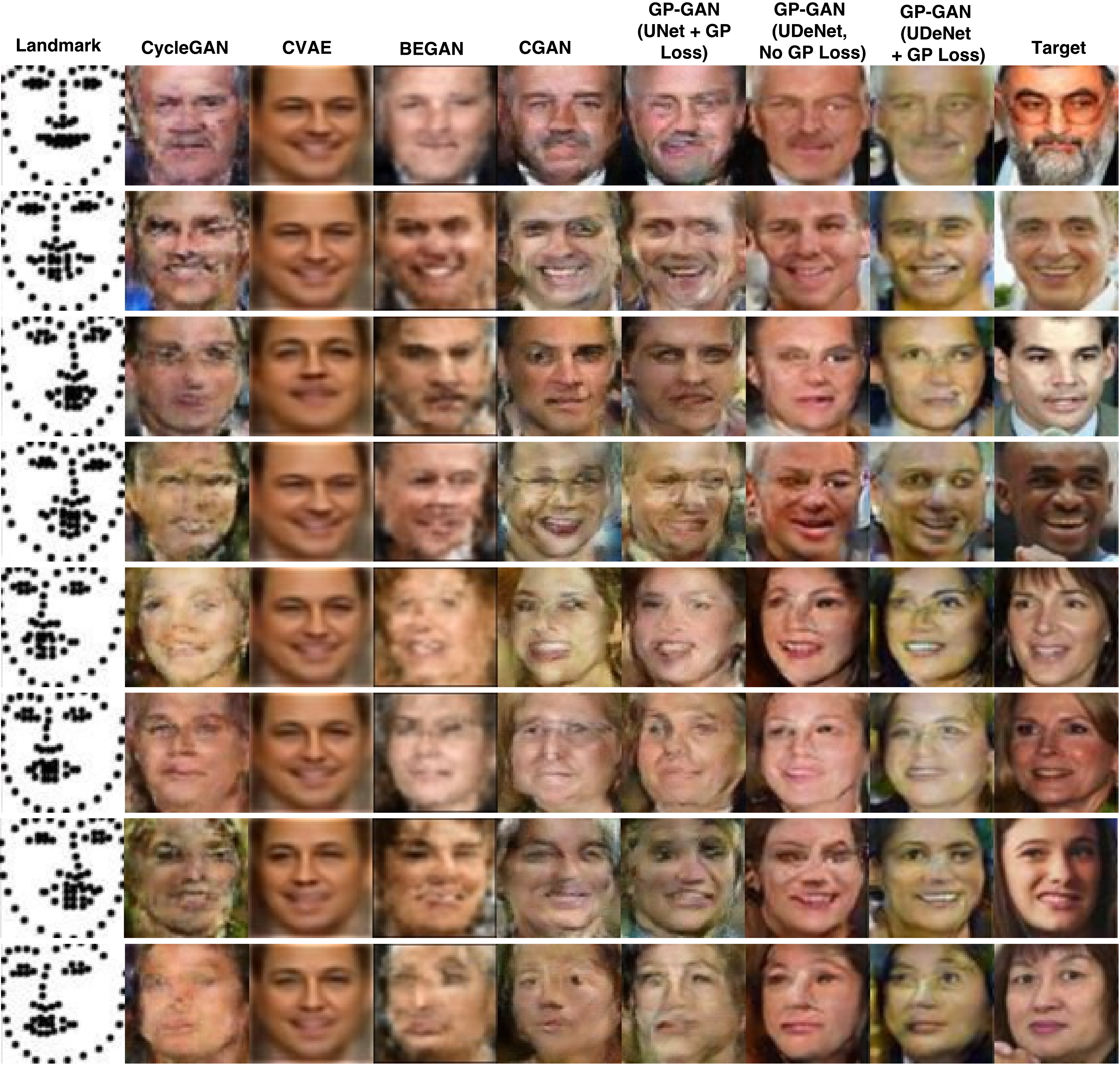}
	\caption{Sample qualitative results of synthesis experiments from LFW dataset. The proposed method GP-GAN (UDeNet + GP Loss) achieves more realistic synthesis compared to the other methods (CycleGAN, CVAE, BEGAN, CGAN) and the baseline methods from the ablation study: GP-GAN (UNet+GP Loss), GP-GAN (UDeNet+ No GP Loss).}
	\label{fig:lfwreconstruction}
\end{figure}

\begin{figure}
	\centering
	\includegraphics[width=1\linewidth]{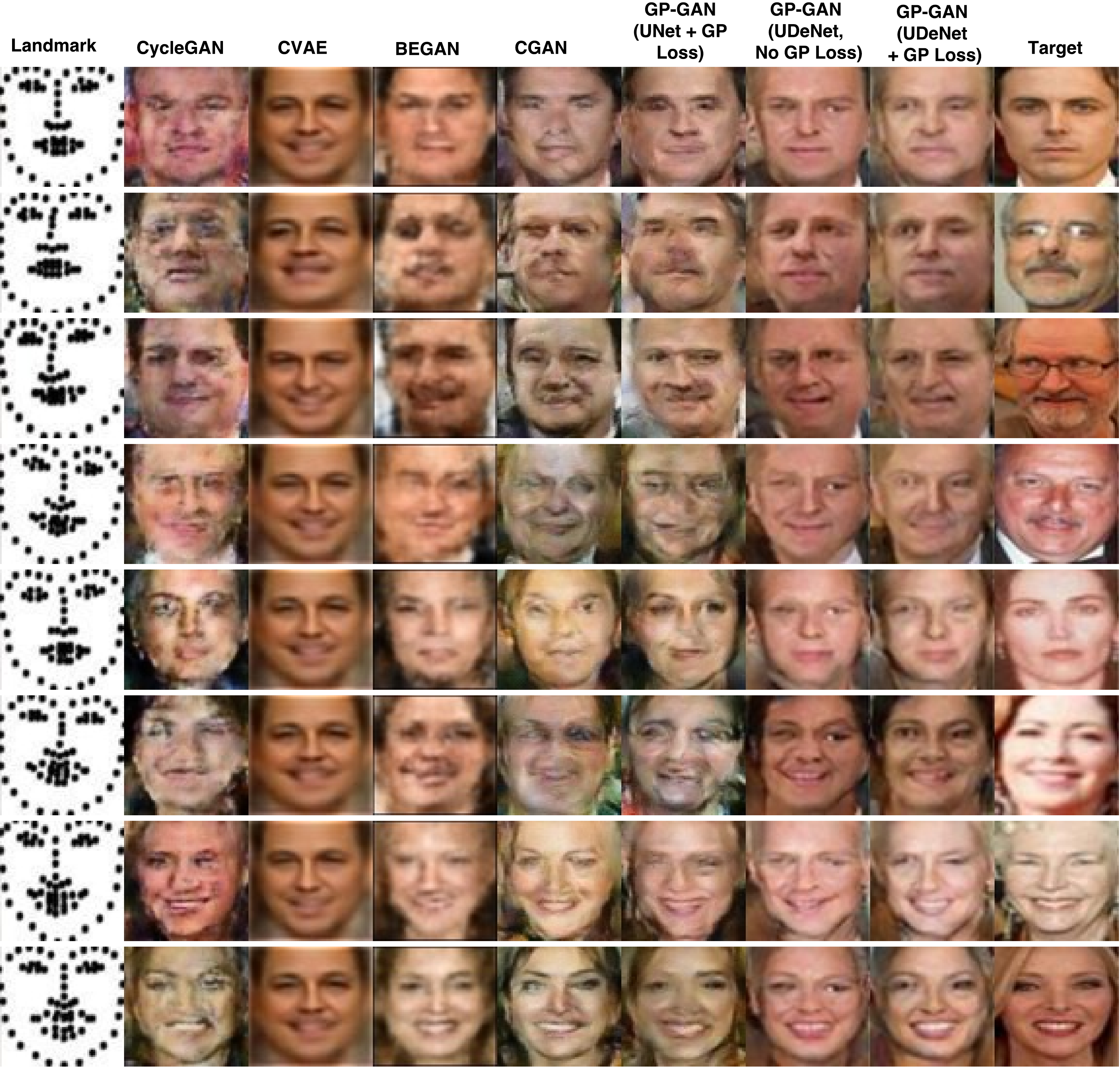}
	\caption{Sample qualitative results of synthesis experiments from CASIA WebFace dataset. The proposed method GP-GAN (UDeNet + GP Loss) achieves more realistic synthesis compared to the other methods (CycleGAN, CVAE, BEGAN, CGAN) and the baseline methods from the ablation study: GP-GAN (UNet+GP Loss), GP-GAN (UDeNet+ No GP Loss). }
	\label{fig:casiareconstruction}
\end{figure}

\noindent\textbf{Gender preserving loss:} Inspired largely by the perceptual loss, we define a gender preserving loss. As indicated by the name, this function measures the error in terms of gender attribute of the synthesized image as compared to that of real image. It is defined as: 
\begin{equation} \label{classification loss}
\nonumber	L_{C} = -\frac{1}{N} \sum_{i}(C(x_i)\log(C(\hat{x}_i))+(1-C(x_i))\log(C(\hat{x}_i)),
\end{equation}
where $C$ represents a pre-trained gender classification network. In this work, $C$ is constructed using the standard VGG-16 network in which, the convolutional layers are retained and the fully connected layers are replaced by a new set of layers as shown in Fig. \ref{fig:framework}. This network is trained by minimizing the standard binary cross entropy error.\\

\noindent\textbf{L\textsc{1} loss:} L\textsc{1} loss measures the reconstruction error between  the synthesized face image and the corresponding real image and is defined as 
\begin{equation} \label{eq:l1loss}
	L_{1} = ||G(\hat{x})-x||_{1}
\end{equation}

\begin{table*}[ht!]
	\caption{Quantitative comparison of gender recognition accuracy (\%) for various methods.}\label{result_accuracy}
	\centering
\vskip-10pt	\resizebox{0.99\textwidth}{!}{%
		\begin{tabular}{|l|c|c|c|c|c|c|c|c|c|}
			\hline  & LM (D) & LM (A) & CycleGAN & CVAE & BEGAN & CGAN & GP-GAN (UNet+GP-Loss) & GP-GAN (UDeNet,No GP Loss) & GP-GAN (UDeNet+GP-Loss)\\ 
			\hline LFW  & 78.0 $\pm$ 1.9 & 79.8 $\pm$ 2.4 & 81.8 $\pm$ 1.1 & 80.3 $\pm$ 2.0  & 84.4 $\pm$ 1.9 & 86.3 $\pm$ 2.5 & 91.1 $\pm$ 1.1 & 91.7 $\pm$ 1.6 & 93.1 $\pm$ 1.2  \\ 
			\hline CASIA  & 61.0 $\pm$ 11.8 & 61.7 $\pm$ 13.6 & 64.8 $\pm$ 3.3 & 62.0 $\pm$ 4.1  & 67.8 $\pm$ 5.0 & 70.4 $\pm$ 5.5 & 73.2 $\pm$3.9 & 76.7 $\pm$ 4.3 & 78.4 $\pm$ 4.1  \\ 
			\hline 
		\end{tabular}
	}

\end{table*}

\section{Experiments and Evaluations}
In this section, experimental settings and evaluation of the proposed method are discussed in detail. We present the qualitative and quantitative results of the synthesis experiment. The quantitative performance is measured using gender recognition rates. Results are compared with four state-of-the-art generative models: Conditional GAN \cite{isola2016image}, Cycle GAN \cite{zhu2017unpaired}, CVAE \cite{sohn2015learning} \cite{yan2016attribute2image} and adopted BEGAN\footnote{https://github.com/taey16/pix2pixBEGAN.pytorch} in addition to two baseline methods (a) GP-GAN using U-Net generator with GP-Loss, and (b) GP-GAN using UDeNet generator without GP-Loss. The baseline comparisons are performed to demonstrate the improvements achieved by the gender preserving loss and UDeNet components. Also, we demonstrate that the use of synthesis using GP-GAN accentuates gender information present in  landmarks by comparing gender recognition rates with methods that directly compute these rates from landmark points \cite{cao2011can}. Furthermore, we conduct an experiment to evaluate the data augmentation capabilities of the synthesis method.

\subsection{Preprocessing and training details}
Prior to preforming these experiments, all images in both datasets are fed through a pre-processing pipeline. First,  MTCNN \cite{zhang2016joint} is employed for detecting face bounding boxes which are further used to crop the faces followed by landmark key point detection using TCDCN algorithm \cite{zhang2014facial}. Pairs of these detected landmarks and faces are used for training the proposed method. Since we consider this problem as an image-to-image translation, the input landmark is encoded using a heatmap (similar to \cite{ma2017pose}) as shown in Fig. \ref{fig:overview} which is a created by imposing a 2D Gaussian with standard deviation of 0.2 at every landmark location on a blank image like could counting work \cite{sindagi2017generating}. Note that the cropped face images are resized to 64$\times$64. 

The proposed network is trained on a single TitanX GPU for approximately 10 hours (200 epochs). A learning rate of $2\times 10^{-4}$ is used for $G$ and $D$. For perceptual network, the input images are resized to a size of $224\times224$. The learning rate is decayed by a factor of $2\times 10^{-6}$ for every epoch after 100 epochs. The weights $\lambda_{A}$, $\lambda_{P}$ and $\lambda_{C}$ are set equal to  $100$, $1$ and $1$, respectively. 

For learning the parameters of the proposed method and baselines, training set from the LFW official deep funneling aligned dataset \cite{huang2007labeled}\cite{Huang2012a} is used. It contains 5749 identities, and 13233 images. The official training, validating and testing View 1 was used for this experiment. After detection and crop procedure, we are left with 3757 images in the training set and 1615 images in the test set. The trained network is evaluated on the LFW test set and a subset of CASIA-Webface dataset \cite{yi2014learning}. The test subset for CASIA-Webface is constructed by randomly selecting 1000 male and 1000 female face images. Note that, in order to demonstrate the generalization performance, the proposed network is trained using only the LFW training set and evaluated on the LFW test set and the CASIA-Webface dataset.

\subsection{Results}

Fig. \ref{fig:lfwreconstruction} and Fig. \ref{fig:casiareconstruction} show sample results of reconstruction using various methods on the LFW and CASIA datasets, respectively.  The landmark image is used as the input for all the methods except CVAE \cite{sohn2015learning} \cite{yan2016attribute2image}. For CVAE, the inputs are original image and normalized landmark locations as the attributes. It can be clearly observed that Conditional GAN \cite{isola2016image}, Cycle GAN \cite{zhu2017unpaired} and BEGAN \cite{berthelot2017began} are unable to reconstruct visually coherent faces. Though CVAE is able to generate visually appropriate faces, they fail to preserve the gender information.  Since their network implements an auto-encoder like architecture and uses pixel-wise Euclidean measure, the output is often blurry, due to which gender classification becomes very difficult. GP-GAN using UDeNet generator without GP-Loss is able to generate perceptually better results as compared to GP-GAN using UNet generator with GP-Loss demonstrating the superior performance obtained using the novel combination of UNet and DenseNet architectures. The proposed method GP-GAN (UDeNet and GP-Loss) outperforms all existing and baseline methods. It may be argued that identity information is lost during the reconstruction process, however, note that the goal of the proposed method is not to capture the exact mapping between landmarks and corresponding faces. Instead, the idea is to explore generation of visually coherent faces from landmark keypoints which can further assist in data augmentation and other tasks. 

As discussed earlier, the quantitative performance is measured in terms of gender recognition rates and it is shown in Table \ref{result_accuracy}. Gender recognition rates for the synthesized are calculated using the LBP features  \cite{ojala2002multiresolution} and a linear SVM classifier that is trained using the LFW training set, whereas the recognition rates for landmarks, LM(D) and LM(A), are calculated using the distance and angle methods described in \cite{cao2011can}. Note that the gender recognition is performed based only on landmark keypoints considering that the corresponding face images are unavailable and hence recent state-of-the-art gender recognition methods cannot be used for comparison as they operate on actual face images rather than only on facial landmarks.  Similar to the observations made using visual comparisons, it can be found from the quantitative results that, gender recognition rates improve in general using the generative models as compared to the landmark-based methods.


With respect to the baseline comparisons, it can be observed that GP-GAN using UDeNet generator without GP-Loss outperforms GP-GAN using UNet generator with GP-Loss in spite of the fact that GP-Loss is not used, thus indicating the effectiveness of UDeNet architecture. Furthermore, the proposed method GP-GAN (UDeNet with GP-Loss) outperforms all existing baseline methods by a large margin in terms of gender recognition rates. This indicates that the proposed synthesis method can be used to generate face images from just facial landmarks while retaining gender information present in these landmarks. 

\begin{figure}
	\centering
	\includegraphics[width=0.8\linewidth]{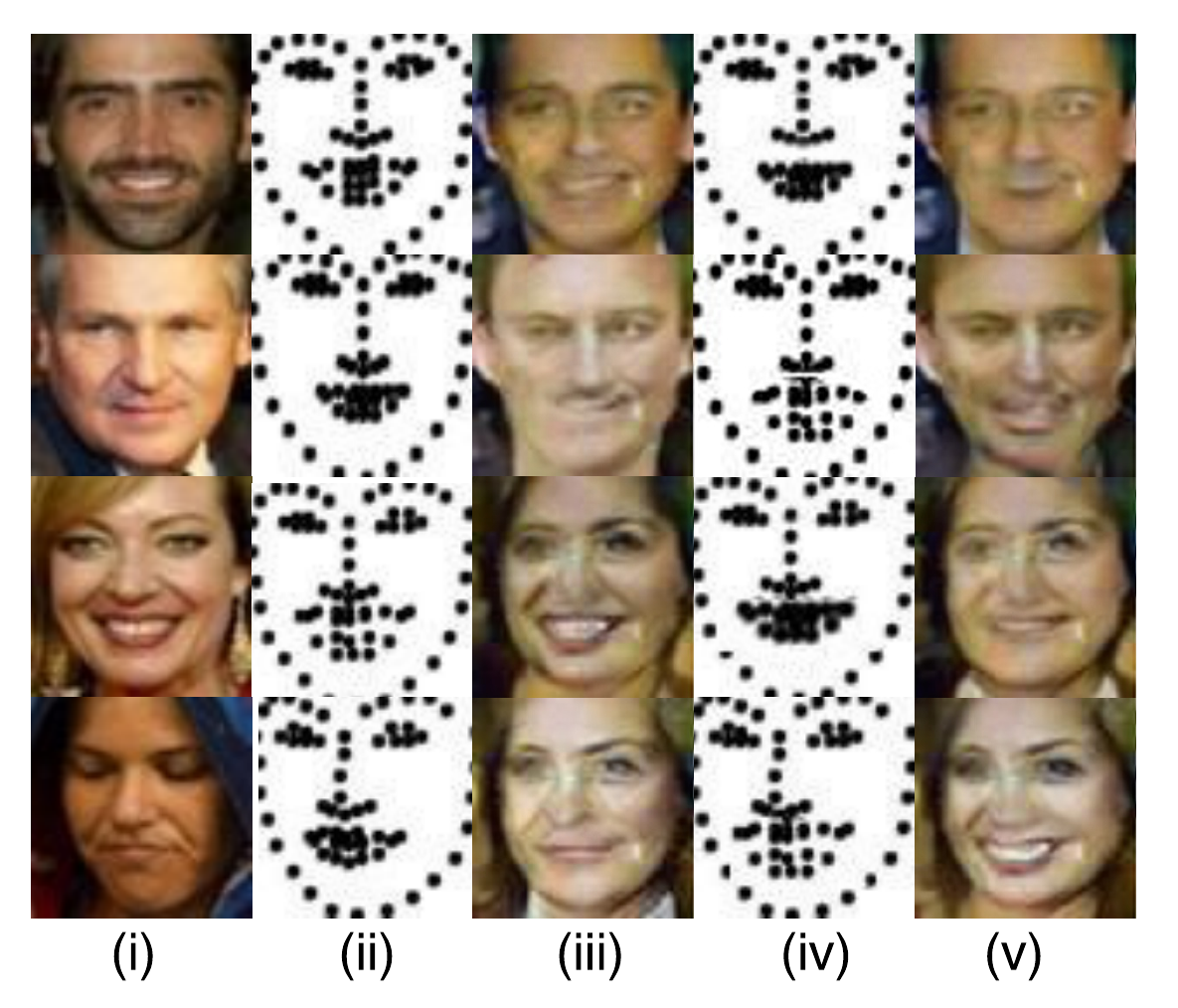}
	\caption{Results of experiment for dataset augmentation where landmark corresponding to a face is modified and used for synthesis. We are able to generate new samples while preserving gender information. (i) Original face image. (ii) Landmark corresponding to original face. (iii) Synthesized face from original landmark. (iv) Landmark obtained after manipulating original landmark. (v) Synthesized face image using manipulated landmark.}
	\label{fig:data_augmentation}
\end{figure}

In addition, we conducted a face synthesis experiment to verify if the proposed method can be used for data augmentation. In this experiment, we manipulate the landmark of a face (for instance, modify mouth open to mouth close) and use this landmark to synthesize a face using generator $G$. Sample results for this experiment are shown in Fig~\ref{fig:data_augmentation}. It can be seen that, the generator $G$ is able to synthesize realistic faces from the modified landmarks while reflecting this modification in the synthesized face. Additionally, the gender attribute is also retained.  Based on these experiments, we can conclude that the proposed method is able successfully generate face samples which can be used for data augmentation for other facial analysis tasks. 

\section{Conclusion}
We explored the problem of synthesizing faces from landmarks points using the recently introduced generative models. The aim of this project was to demonstrate that information (especially gender) present in the landmark keypoints can be accentuated using synthesis models while generating realistic images. The proposed network is based on the generative adversarial networks and is guided by perceptual loss and a novel gender preserving loss. Further, we propose a novel generator based on UNet and DenseNet architectures. Evaluations are performed on two popular datasets, LFW and CASIA-Webface, and the results are compared with recent state-of-the-art generative methods. It is clearly demonstrated that the proposed method achieves significant improvements in terms of visual quality and gender recognition. Additionally, we conducted a face synthesis experiment to demonstrate that the proposed generative method can be used as a data augmentation technique.

\section*{ACKNOWLEDGMENTS}
This research is based upon work supported by the Office of the Director
of  National  Intelligence  (ODNI),  Intelligence  Advanced  Research  Projects
Activity  (IARPA),  via  IARPA  R\&D  Contract  No.  2014-14071600012.  The
views and conclusions contained herein are those of the authors and should not
be interpreted as necessarily representing the official policies or endorsements,
either  expressed  or  implied,  of  the  ODNI,  IARPA,  or  the  U.S.  Government.
The  U.S.  Government  is  authorized  to  reproduce  and  distribute  reprints  for
Governmental purposes notwithstanding any copyright annotation thereon.   We like to thank He Zhang for his insightful discussion on this topic.  



{\small
	\bibliographystyle{ieee}
	\bibliography{egbib}

\begin{thebibliography}{10}\itemsep=-1pt

\bibitem{asthana2013robust}
A.~Asthana, S.~Zafeiriou, S.~Cheng, and M.~Pantic.
\newblock Robust discriminative response map fitting with constrained local
  models.
\newblock In {\em Proceedings of the IEEE Conference on Computer Vision and
  Pattern Recognition}, pages 3444--3451, 2013.

\bibitem{berthelot2017began}
D.~Berthelot, T.~Schumm, and L.~Metz.
\newblock Began: Boundary equilibrium generative adversarial networks.
\newblock {\em arXiv preprint arXiv:1703.10717}, 2017.

\bibitem{bousmalis2016unsupervised}
K.~Bousmalis, N.~Silberman, D.~Dohan, D.~Erhan, and D.~Krishnan.
\newblock Unsupervised pixel-level domain adaptation with generative
  adversarial networks.
\newblock {\em arXiv preprint arXiv:1612.05424}, 2016.

\bibitem{cao2011can}
D.~Cao, C.~Chen, M.~Piccirilli, D.~Adjeroh, T.~Bourlai, and A.~Ross.
\newblock Can facial metrology predict gender?
\newblock In {\em Biometrics (IJCB), 2011 International Joint Conference on},
  pages 1--8. IEEE, 2011.

\bibitem{JC_CNN}
J.~C. Chen, V.~M. Patel, and R.~Chellappa.
\newblock Unconstrained face verification using deep cnn features.
\newblock In {\em 2016 IEEE Winter Conference on Applications of Computer
  Vision (WACV)}, pages 1--9, March 2016.

\bibitem{di2017face}
X.~Di and V.~M. Patel.
\newblock Face synthesis from visual attributes via sketch using conditional
  vaes and gans.
\newblock {\em arXiv preprint arXiv:1801.00077}, 2017.

\bibitem{goodfelllow2014generative}
I.~Goodfellow, J.~Pouget-Abadie, M.~Mirza, B.~Xu, D.~Warde-Farley, S.~Ozair,
  A.~Courville, and Y.~Bengio.
\newblock Generative adversarial nets.
\newblock In Z.~Ghahramani, M.~Welling, C.~Cortes, N.~D. Lawrence, and K.~Q.
  Weinberger, editors, {\em Advances in Neural Information Processing Systems
  27}, pages 2672--2680. Curran Associates, Inc., 2014.

\bibitem{huang2017densely}
G.~Huang, Z.~Liu, L.~van~der Maaten, and K.~Q. Weinberger.
\newblock Densely connected convolutional networks.
\newblock In {\em Proceedings of the IEEE Conference on Computer Vision and
  Pattern Recognition}, 2017.

\bibitem{Huang2012a}
G.~B. Huang, M.~Mattar, H.~Lee, and E.~Learned-Miller.
\newblock Learning to align from scratch.
\newblock In {\em NIPS}, 2012.

\bibitem{huang2007labeled}
G.~B. Huang, M.~Ramesh, T.~Berg, and E.~Learned-Miller.
\newblock Labeled faces in the wild: A database for studying face recognition
  in unconstrained environments.
\newblock Technical report.

\bibitem{isola2016image}
P.~Isola, J.-Y. Zhu, T.~Zhou, and A.~A. Efros.
\newblock Image-to-image translation with conditional adversarial networks.
\newblock {\em arXiv preprint arXiv:1611.07004}, 2016.

\bibitem{johnson2016perceptual}
J.~Johnson, A.~Alahi, and L.~Fei-Fei.
\newblock Perceptual losses for real-time style transfer and super-resolution.
\newblock In {\em European Conference on Computer Vision}, pages 694--711.
  Springer, 2016.

\bibitem{kingma2013auto}
D.~P. Kingma and M.~Welling.
\newblock Auto-encoding variational bayes.
\newblock {\em arXiv preprint arXiv:1312.6114}, 2013.

\bibitem{kumar2016face}
A.~Kumar, R.~Ranjan, V.~Patel, and R.~Chellappa.
\newblock Face alignment by local deep descriptor regression.
\newblock {\em arXiv preprint arXiv:1601.07950}, 2016.

\bibitem{ma2017pose}
L.~Ma, X.~Jia, Q.~Sun, B.~Schiele, T.~Tuytelaars, and L.~Van~Gool.
\newblock Pose guided person image generation.
\newblock In {\em NIPS}, 2017.

\bibitem{mao2016image}
X.-J. Mao, C.~Shen, and Y.-B. Yang.
\newblock Image denoising using very deep fully convolutional encoder-decoder
  networks with symmetric skip connections.
\newblock {\em arXiv preprint}, 2016.

\bibitem{mirza2014conditional}
M.~Mirza and S.~Osindero.
\newblock Conditional generative adversarial nets.
\newblock {\em arXiv preprint arXiv:1411.1784}, 2014.

\bibitem{ojala2002multiresolution}
T.~Ojala, M.~Pietikainen, and T.~Maenpaa.
\newblock Multiresolution gray-scale and rotation invariant texture
  classification with local binary patterns.
\newblock {\em IEEE Transactions on pattern analysis and machine intelligence},
  24(7):971--987, 2002.

\bibitem{HyperFace}
R.~Ranjan, V.~M. Patel, and R.~Chellappa.
\newblock Hyperface: A deep multi-task learning framework for face detection,
  landmark localization, pose estimation, and gender recognition.
\newblock {\em IEEE Transactions on Pattern Analysis and Machine Intelligence},
  pages 1--1, 2017.

\bibitem{DeepFace_SPM2018}
R.~Ranjan, S.~Sankaranarayanan, A.~Bansal, N.~Bodla, J.~C. Chen, V.~M. Patel,
  C.~D. Castillo, and R.~Chellappa.
\newblock Deep learning for understanding faces: Machines may be just as good,
  or better, than humans.
\newblock {\em IEEE Signal Processing Magazine}, 35(1):66--83, Jan 2018.

\bibitem{reed2016generative}
S.~Reed, Z.~Akata, X.~Yan, L.~Logeswaran, B.~Schiele, and H.~Lee.
\newblock Generative adversarial text-to-image synthesis.
\newblock In {\em Proceedings of The 33rd International Conference on Machine
  Learning}, 2016.

\bibitem{rezende2014stochastic}
D.~J. Rezende, S.~Mohamed, and D.~Wierstra.
\newblock Stochastic backpropagation and approximate inference in deep
  generative models.
\newblock {\em arXiv preprint arXiv:1401.4082}, 2014.

\bibitem{ronneberger2015u}
O.~Ronneberger, P.~Fischer, and T.~Brox.
\newblock U-net: Convolutional networks for biomedical image segmentation.
\newblock In {\em International Conference on Medical Image Computing and
  Computer-Assisted Intervention}, pages 234--241. Springer, 2015.

\bibitem{saragih2011deformable}
J.~M. Saragih, S.~Lucey, and J.~F. Cohn.
\newblock Deformable model fitting by regularized landmark mean-shift.
\newblock {\em International Journal of Computer Vision}, 91(2):200--215, 2011.

\bibitem{simonyan2014very}
K.~Simonyan and A.~Zisserman.
\newblock Very deep convolutional networks for large-scale image recognition.
\newblock {\em arXiv preprint arXiv:1409.1556}, 2014.

\bibitem{sindagi2017generating}
V.~A. Sindagi and V.~M. Patel.
\newblock Generating high-quality crowd density maps using contextual pyramid
  cnns.
\newblock In {\em IEEE International Conference on Computer Vision}, 2017.

\bibitem{sohn2015learning}
K.~Sohn, H.~Lee, and X.~Yan.
\newblock Learning structured output representation using deep conditional
  generative models.
\newblock In {\em Advances in Neural Information Processing Systems}, pages
  3483--3491, 2015.

\bibitem{Qianru2018HeadInpainting}
Q.~Sun, L.~Ma, S.~J. Oh, L.~V. Gool, B.~Schiele, and M.~Fritz.
\newblock Natural and effective obfuscation by head inpainting.
\newblock In {\em CVPR}, 2018.

\bibitem{sun2013deep}
Y.~Sun, X.~Wang, and X.~Tang.
\newblock Deep convolutional network cascade for facial point detection.
\newblock In {\em Proceedings of the IEEE conference on computer vision and
  pattern recognition}, pages 3476--3483, 2013.

\bibitem{taheri2011towards}
S.~Taheri, P.~Turaga, and R.~Chellappa.
\newblock Towards view-invariant expression analysis using analytic shape
  manifolds.
\newblock In {\em Automatic Face \& Gesture Recognition and Workshops (FG
  2011), 2011 IEEE International Conference on}, pages 306--313. IEEE, 2011.

\bibitem{valstar2010facial}
M.~Valstar, B.~Martinez, X.~Binefa, and M.~Pantic.
\newblock Facial point detection using boosted regression and graph models.
\newblock In {\em Computer Vision and Pattern Recognition (CVPR), 2010 IEEE
  Conference on}, pages 2729--2736. IEEE, 2010.

\bibitem{lidan2017highquality}
L.~Wang, V.~A. Sindagi, and V.~M. Patel.
\newblock High-quality facial photo-sketch synthesis using multi-adversarial
  networks.
\newblock {\em CoRR}, abs/1710.10182, 2017.

\bibitem{wei2018UniqueSmile}
W.~Wang, X.~Alameda{-}Pineda, D.~Xu, E.~Ricci, and N.~Sebe.
\newblock Every smile is unique: Landmark-guided diverse smile generation.
\newblock In {\em CVPR}, 2018.

\bibitem{wu2012age}
T.~Wu, P.~Turaga, and R.~Chellappa.
\newblock Age estimation and face verification across aging using landmarks.
\newblock {\em IEEE Transactions on Information Forensics and Security},
  7(6):1780--1788, 2012.

\bibitem{yan2016attribute2image}
X.~Yan, J.~Yang, K.~Sohn, and H.~Lee.
\newblock Attribute2image: Conditional image generation from visual attributes.
\newblock In {\em European Conference on Computer Vision}, pages 776--791.
  Springer, 2016.

\bibitem{yi2014learning}
D.~Yi, Z.~Lei, S.~Liao, and S.~Z. Li.
\newblock Learning face representation from scratch.
\newblock {\em arXiv preprint arXiv:1411.7923}, 2014.

\bibitem{zhang2018densely}
H.~Zhang and V.~M. Patel.
\newblock Densely connected pyramid dehazing network.
\newblock {\em arXiv preprint arXiv:1803.08396}, 2018.

\bibitem{zhang2017image}
H.~Zhang, V.~Sindagi, and V.~M. Patel.
\newblock Image de-raining using a conditional generative adversarial network.
\newblock {\em arXiv preprint arXiv:1701.05957}, 2017.

\bibitem{zhang2017joint}
H.~Zhang, V.~Sindagi, and V.~M. Patel.
\newblock Joint transmission map estimation and dehazing using deep networks.
\newblock {\em arXiv preprint arXiv:1708.00581}, 2017.

\bibitem{zhang2016joint}
K.~Zhang, Z.~Zhang, Z.~Li, and Y.~Qiao.
\newblock Joint face detection and alignment using multitask cascaded
  convolutional networks.
\newblock {\em IEEE Signal Processing Letters}, 23(10):1499--1503, 2016.

\bibitem{zhang2014facial}
Z.~Zhang, P.~Luo, C.~C. Loy, and X.~Tang.
\newblock Facial landmark detection by deep multi-task learning.
\newblock In {\em ECCV}, pages 94--108. Springer, 2014.

\bibitem{zhao2016energy}
J.~Zhao, M.~Mathieu, and Y.~LeCun.
\newblock Energy-based generative adversarial network.
\newblock {\em arXiv preprint arXiv:1609.03126}, 2016.

\bibitem{zhu2017unpaired}
J.-Y. Zhu, T.~Park, P.~Isola, and A.~A. Efros.
\newblock Unpaired image-to-image translation using cycle-consistent
  adversarial networks.
\newblock {\em arXiv preprint arXiv:1703.10593}, 2017.

\end{thebibliography}
}

\end{document}